\pdfoutput=1

\documentclass[10pt, a4paper]{article}

\usepackage{lrec-coling2024} 

\usepackage[T1]{fontenc}

\usepackage[utf8]{inputenc}

\usepackage{microtype}
\usepackage{booktabs}
\usepackage{lipsum}
\usepackage{paralist}
\usepackage{annotates}
\usepackage{multirow}
\usepackage{amssymb}
\usepackage{graphicx}
\usepackage{tabularx}
\usepackage{rotating}
\usepackage{pifont}

\newcolumntype{C}{>{\centering\arraybackslash}X}

\title{Towards Understanding the Relationship  between \\ In-context Learning and Compositional Generalization}

\name{Sungjun Han$^{\ast}$, Sebastian Padó$^{\dagger}$}

 \address{$^{\ast}$42dot Inc., 
          \texttt{sungjun.han@42dot.ai} \\
          $^{\dagger}$IMS, University of Stuttgart, \texttt{pado@ims.uni-stuttgart.de} }

\abstract{
  According to the principle of \textit{compositional generalization},
  the meaning of a complex expression can be understood as a function of the meaning of its parts and of how they are combined. This principle is crucial for human language processing and also, arguably, for NLP models in the face of out-of-distribution data.
  However, many neural network models, including Transformers, have
  been shown to struggle with compositional generalization. In this
  paper, we hypothesize that forcing models to \textit{in-context
    learn} can provide an inductive bias to promote compositional
  generalization. To test this hypothesis, we train a causal
  Transformer in a setting that renders 'ordinary' learning very
  difficult: we present it with different orderings of the training
  instance and shuffle instance labels. This corresponds to training
  the model on all possible few-shot learning problems attainable from
  the dataset. The model can solve the task, however, by utilizing
  earlier examples to generalize to later ones -- i.e., in-context
  learning.  
  In evaluations on the datasets, SCAN, COGS, and GeoQuery, models trained in this
  manner indeed show improved compositional generalization. This
  indicates the usefulness of in-context learning problems as
  an inductive bias for generalization.
 \\ \newline }

\begin{document}
\maketitleabstract
\bibliographystyle{lrec-coling2024-natbib}

\section{Introduction}

As humans, we have the ability to combine atomic parts in reoccurring
structures in novel manners \citep{fodor_connectionism_nodate}. This
ability, known as \textit{compositional generalization}, is an
important aspect of human language processing, affording us with an
"infinite use of finite means" \citep{chomsky1965}. For example, when
we understand the meaning of the predicate \textit{dax} in phrases
such as ``I can dax'' and ``dax twice'', we can also understand
phrases such as ``dax voluntarily'' or ``must dax''.



In contrast, many modern deep neural architectures struggle with
compositional generalization \citep{baroni_linguistic_2020, lake_scan,
  hupkes_compositionality_2020, kim_cogs_2020,
  keysers_measuring_2020}. While they excel at making predictions for
test sets similarly distributed to the training set (i.e.,
\textit{in-distribution}), their performances significantly decrease
when generalizing to test distributions that are differently
structured (i.e., \textit{out-of-distribution}) even if they contain
the same set of atoms.


We believe that standard models lack an inductive bias towards
acquiring compositional representation, which arises from the
independent parallel processing of examples in mini-batches. In most
mini-batches, the models do not have \textit{explicit} access to a
sufficient number of instances of the atoms to make it worthwhile to
learn compositionally generalizable representations for the
atoms. Contrast this with symbolic accounts of compositional
generalization, e.g., in the shape of \textit{case-based reasoning}
\citep{leake_case_based}, where prediction can always rely on the
availability of a sufficient number of relevant examples in
memory. Thus, the ability to understand “dax thrice” from "dax twice"
can be thought of as a generalization of relevant past uses of
"thrice" in memory, such as “eat thrice”, combined with the use of
"dax" in "dax twice".




If this is true, then compositional generalization should be
encouraged by forcing models to \textit{in-context learn}
\citep{brown2020language, chowdhery2022palm} -- that is, forcing them
to generalize to new examples conditioned on a few demonstrations of
input-output mappings provided in the model's context (or memory)
without parameter updates.
In-context learning forces the model to compute in the forward pass how the past examples provided the context can be utilized in a novel manner for the later examples. We observe that it is the same mechanism that supports the learning of compositionally generalizable input-output mappings.

The intuition behind our study is aligned with theoretical studies
that explain in-context learning \citep{ortega_meta-learning_2019,
  xie_explanation_2022} as an implicit Bayesian inference, where the
model learns to approximate the latent parameters. However, it is yet
unclear empirically how compositional generalization and in-context
learning are related. On the one hand, the reported improvement in
compositional generalization for the large Transformer-based LLMs
\citep{zhou2023leasttomost, hosseini2022compositional} with
\textit{emergent} in-context learning ability seem to point to an
underlying relationship. On the other hand, our understanding is
limited by (a) the lack of control over the training data in these
studies and (b) the uncertainty regarding how much of the inductive
biases implicit in the prompting methods contribute to the
improvement. Indeed, \citet{hosseini2022compositional} and
\citet{qiu-etal-2022-evaluating} report that only some in-context
learning LLMs can compositionally
generalize and only as they scale up.




As implementation, we utilize a \textit{meta-learning}
\citep{schmidhuber_meta_learning, bengio_meta_learning,
hochreiter_meta_learning, duan_rl2_2016,ortega_meta-learning_2019}
regime to explicitly incentivize in-context learning for a causal
Transformer \citep{vaswani, radford2019language} with a language
modelling objective. We train \textit{from scratch} to eliminate the possible confounders introduced from pre-training in studying the relationship. Each task of our
\textit{meta}-task distribution is one possible linear ordering of
input-output pairs of the training dataset formed into a single
sequence via concatenation. This trains the model
on all possible few-shot in-context learning problems attainable from
the dataset. In order to discourage the model from relying on memorization, we also shuffle the labels.
At prediction time, we condition the inference on the
test examples on randomly sampled training mappings, maintaining the
\textit{zero-shot} prediction setting. We evaluate our approach on three
widely used datasets targeting specifically compositional generalization, namely
SCAN \citep{lake_scan, keysers_measuring_2020}, COGS \citep{kim_cogs_2020}, and GeoQuery \citep{geoquery, shaw-etal-2021-compositional}.
Our contributions are:






\begin{enumerate}
    \item We empirically study the relationship between in-context learning and compositional generalization through a novel meta-learning training regime that incentivizes in-context learning on established compositional generalization datasets along with a corresponding evaluation regime that maintains a zero-shot prediction setting.
    \item We show that a causal Transformer trained through meta-in-context learning from scratch without any pretraining exhibits a significant improvement in performance on compositional generalization compared to the models without meta-learning.
    \item We demonstrate several connections between in-context learning and compositional generalization through ablations: More in-context learning problems lead to better compositional generalization (Exp. 2); trained models are indeed generalizing through in-context learning in informative contexts (Exp. 3); the success of in-context learning depends on the absence of memorization (Exp.\ 4);  pre-trained models have a better prior for in-context learning and can also
      benefit from meta-learning (Exp. 5).
\end{enumerate}
%
%
%
%
\paragraph{Plan of the paper}  \S \ref{s:related_works} introduces important background concepts and reviews notable related works. \S \ref{s:method} presents our meta-learning regime in detail. \S \ref{s:experimental_setup} provides information on experimental setup, followed by the experimental results in \S \ref{s:results}. \S \ref{s:conclusion} concludes the paper along with future directions.



\section{Related Work} \label{s:related_works}


\subsection{Compositional Generalization}

The difficulties of neural networks in compositional generalization have been identified by many studies. In the following, we focus on studies on unimodal language data.

Notable text-to-text benchmarks include SCAN \citep{lake_scan}, PCFG \citep{hupkes_compositionality_2020}, COGS \citep{kim_cogs_2020}, and CFQ \citep{keysers_measuring_2020}. These datasets generate a single data distribution which are split into a train and test set in a systematic manner, attempting to capture the notion of \textit{systematicity} or/and \textit{productivity} \citep{fodor_connectionism_nodate, hupkes_compositionality_2020}. The former refers to the ability to recombine parts in a novel manner, and the latter to the ability to recursively combine known structures. 
 Hence, if a model learning from the train set can find a compositional solution, it can be successful on the test as the same data generative process underlie the two.

Many studies have proposed different inductive biases to promote compositionality. They include new deep learning architectures structurally constraining the ways the inputs are processed and represented \citep{li_compositional_2019,russin_compositional_2019, Gordon2020Permutation, bergen_systematic_2021}, providing additional supervisory signals \citep{ jiang-bansal-2021-inducing}, data augmentation \citep{andreas-2020-good, guo_revisiting_2020, akyürek2021learning, qiu-etal-2022-improving, li-etal-2023-learning}, and hybrid symbolic reasoning approaches \citep{nye_learning_2020, liu_2020_lane, guo_poset_2020}. These approaches have shown to improve compositional generalization. However, they often require prior knowledge of the dataset, and their scalability to bigger and more general datasets is uncertain.

Following these concerns, some studies have constrained their investigations to the popular neural sequences model such as Transformers \citep{ontanon-etal-2022-making, csordas_devil_2022}, finding that their compositional generalization capacity can be improved with the available variants (e.g. relative positional encoding \citep{dai_transformer-xl_2019} or tying the layers \citep{dehghani2019universal}). \citet{patel-etal-2022-revisiting} showed that popular architectures including Transformers can be improved by increasing diversity in the data distribution. Our work follows this line of research by studying how a better inductive bias can be provided without a major change in the Transformer architecture.

\subsection{Meta-learning}
Meta-learning \citep{bengio_meta_learning, schmidhuber_meta_learning} aims to enable machine learning models to learn how to learn by exposing them to a \textit{distribution of tasks} where a model can improve from experience. The tasks are selected to be similarly structured but differ in details such that it is profitable for the model to find a generalizable solution rather than memorizing individual answers. Our work follows the line of work known as \textit{memory-based meta-learning} or \textit{meta-in-context learning} \citep{hochreiter_meta_learning, santoro_one-shot_2016, duan_rl2_2016, wang_learning_2017,  ortega_meta-learning_2019}, which incentivizes the model to learn to in-context learn by training on a task distribution of \textit{sequences} of input-output mappings.


Meta-learning was applied to various tasks in language processing such as cross-lingual transfer \citep{gu2018metalearning}, question answering \citep{nooralahzadeh2020zeroshot}, and domain adaption  \citep{qian-yu-2019-domain}. However, it has rarely found application in semantic processing. The challenge arises from the difficulty of not knowing beforehand the relevance of specific examples, which makes it difficult to construct the task distribution with the right inductive bias for compositional generalization.
%
\citet{lake_compositional_2019} evaded this problem by using the
ground truth grammar of the data distribution. This allowed them to
permute only the input-output mappings of the primitives, which
was shown to improve compositional generalization. \citet{conklin_meta-learning_2021} used model-agnostic meta learning (MAML)  \citep{finn-maml} as an auxiliary loss for supervised learning. In this approach, a single gradient step is taken on one set of support examples and the auxiliary loss is accrued by how well the updated model performs on another structurally similar set. This loss is back-propagated through the gradient optimization step all the way back to the model weights. The proposed method alleviated the problem of selecting support examples during evaluation, but the approach still relied on ground truth structural knowledge. 




\subsection{In-context Learning}
A long line of work attempts to understand the property of in-context learning, especially related to their ability to generalize to out-of-distribution data. A number of studies has shown that in-context learning in LLMs can be utilized for compositional generalization using specific prompting methods \citep{zhou2023leasttomost, wei2022chain, fu2023complexitybased} or by scaling up the model \citep{hosseini2022compositional, qiu-etal-2022-evaluating}. One recent work by \citet{an-etal-2023-context} has also investigated how the in-context learning ability of LLMs can be improved by selecting better demonstrations with relevant linguistic structure. As explained above, the in-context learning ability in these models was also analyzed theoretically, and the driving force was found to be latent text properties that heavily affects token distributions \citep{xie_explanation_2022}. 

Some works have studied the effects of further meta-training LLMs for in-context learning \citep{chen-etal-2022-meta, min-etal-2022-metaicl} and found the tuned models to perform better than the base LLM. Our work is closer to the studies that train Transformers from \textit{scratch} instead of looking at LLMs. \citet{chan_data_2022} showed that the emergence of in-context learning to depend on the informativeness of contexts. \citet{garg_2022_linear_icl} showed that Transformers are able to in-context learn simple functions and generalize to out-of-distribution samples, and \citet{kirsch2022generalpurpose} extended its study to in-context learning arbitrary image-label mappings.




\section{Methods} \label{s:method}

We now introduce a meta-learning regime that can be generally applied to a sequence to sequence dataset consisting of input-output sequence pairs. The main goal is how to construct a meta task-distribution with the right inductive bias for compositional generalization. The key idea is the inductive bias created by \textit{online} learning an entire dataset: The model observes each example in the dataset only once and sequentially one after the other. When learning on such a linear ordering of examples (i.e., trajectory), the model cannot memorize and needs to successfully store and represent the past examples to generalize to the future examples.

Since there is no inherent order between the examples in a sequence to sequence dataset, different linear orderings of the dataset pose different generalization problems for a model. However, no matter which ordering we choose, the structure behind each trajectory remains invariant as it is governed by the same set of latent parameters. Hence, when meta-learning on such a task distribution, a model has a chance of approximating the underlying structure of the dataset. Note that this way of constructing the task distribution do not require any prior knowledge of the dataset, in contrast to the earlier approaches \citep{lake_scan, conklin_meta-learning_2021}.



 \begin{figure*}[tb]
  \centering
  \includegraphics[width=1\textwidth]{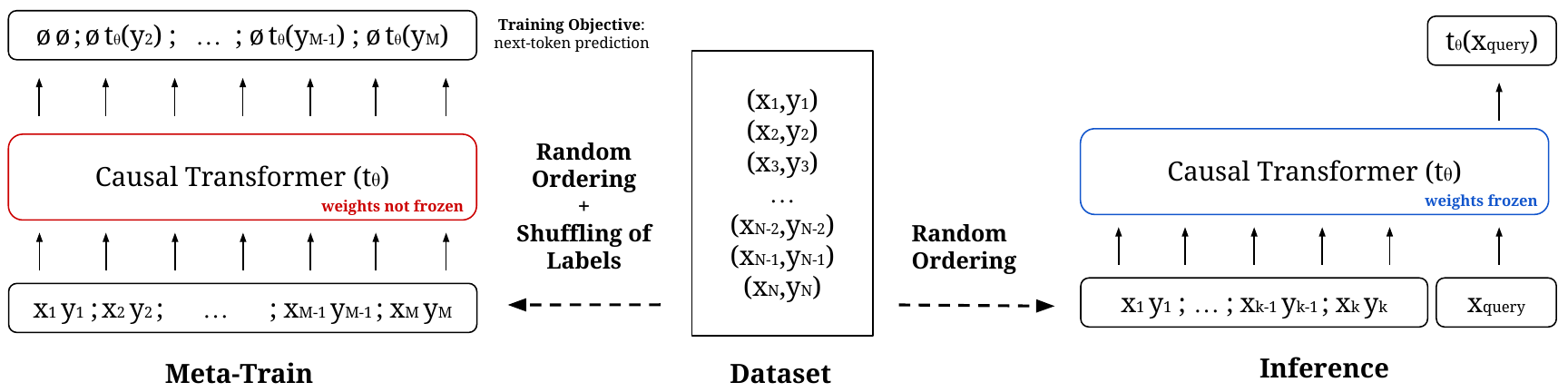}
    \caption{Illustration of our meta-in-context learning framework. (Left) We build our meta-task distribution by sampling random linear orderings from a sequence to sequence dataset and concatenating the input-output mappings (i.e., $(x_i, y_i)$). We optionally shuffle the labels to eliminate memorization and keep only $M$ examples. A causal Transformer ($t_\theta$) is trained with these concatenated results for next-token prediction, only predicting for the outputs. $\phi$ refers to the pad-token. (Right) At inference, we freeze the weights and randomly sample $k < M$ train examples to use as a context in predicting the test query $x_{query}$. }
  \label{fig:illustration}
\end{figure*}

\subsection{Meta-training}
Given a sequence to sequence dataset $D=\{({\mathbf{x}^{(i)}}, {\mathbf{y}^{(i)}})\}_{i=1}^N$ with a vocabulary $\mathrm{V}$, we form the task distribution $P(\tau)$ for meta-learning, where each task $\tau$ is one possible linear ordering of the dataset $({\mathbf{x}^{(1)}}, {\mathbf{y}^{(1)}}), \dots ({\mathbf{x}^{(N)}},{\mathbf{y}^{(N)}})$. 
We feed this to the model as a concatenation $\tau = [{\mathbf{x}^{(1)}};{\mathbf{y}^{(1)}};\dots;{\mathbf{x}^{(N)}};{\mathbf{y}^{(N)}}]$ using two delimiter tokens, one to distinguish the inputs from the outputs and the other to separate the sequence elements. We assume a uniform distribution for $P(\tau)$.

A limitation of this approach is the possibility of memorization as each example occurs many times \textit{across} different trajectories, although it occurs only once \textit{within} each trajectory. Hence, a model might learn to ignore the context and memorize the examples, which is especially true for small datasets. To counteract this danger, we
randomly shuffle the labels of the vocabulary $\mathrm{V}$. For example, given a dataset $\{(jump, J), (run\:twice, R\:R)\}$, we can create an alternate version of the dataset $\{(jump, R), (run\:twice, J\:J)\}$ by the shuffling all instances of $J$ with $R$. This results in more than one output label to be assigned to each input token throughout training and can prevent memorization.

Formally, we train a model $M_\theta$ given a linear ordering sequence of tokens upto and including the $i$-th token $\tau[i]$ to predict the next token $f(\tau[i]) = \tau[i+1]$ if the $i$-th token belongs to the output and a pad token $f(\tau[i]) = \phi $ if it belongs to the input or the first output of the sequence.  The objective is to minimize the expected loss over all possible orderings, optionally shuffling the labels\footnote{For clarity, we do not
formalize the label shuffling in the following equation.}:
\begin{equation}
    \min_\theta \mathrm{E}_{\tau \sim P(\tau)}[\sum_{i=1}^{|\tau|}\frac{1}{|\tau|}\ell(M_\theta(\tau[i]),f(\tau[i]))]
\end{equation}
where $\ell(\cdot,\cdot)$ is the cross-entropy loss function. Figure \ref{fig:illustration} (left half) illustrates the  training procedure.

This objective can be interpreted as training a model on all possible few-shot learning problems attainable from the dataset. Hence, any specific way of defining the meta task-distribution is a subset of our distribution. The strength of the injected inductive bias for compositional generalization is limited to the kinds of generalization problems inherent in each dataset.
A final practical problem is that for most datasets, the set of all
shuffled variants does not fit into GPU memory. Hence, we fix a
certain roll-out length $M < N$ to limit each sequence $\tau$ to
consist of $M$ input-output pairs. Note that as we make $M$ smaller,
the number of distinct tasks in the task distribution decreases. We
investigate the impact of $M$ in Exp.\ 2 below.

\paragraph{Underlying Neural Network.}
This meta-training can be applied to any neural network model with memory. However, the use of an autoregressive model is very advantageous: In such a model, a single trajectory consisting of $N$ concatenated input-output mappings can be combined with causal masking to yield $k-1$ few-shot learning problems in one go. In a bidirectional model, in contrast, 
one needs to provide $k-1$ different problems separately. Hence, we adopt the causal Transformer \citep{vaswani, radford2019language}.

\subsection{Inference}
Compositional generalization datasets are designed as a zero-shot
generalization task. This means that a model is required to
generalize to the test examples only by using the train
examples. 
We randomly sample a
\textit{training} trajectory of length $k < M$ to condition the
inference for each test input $\mathbf{x}_q$. Although we
cannot guarantee the relevance of every sample, the model can still
\textit{choose} among these samples through the attention mechanism,
analogously to case-based reasoning \cite{leake_case_based}. See Figure \ref{fig:illustration} (right half) for the illustration of our inference method. We investigate the implications of the
choice of $k$ in Exp.\ 3 below.

\section{Experimental Setup} \label{s:experimental_setup}



\subsection{Datasets}
\paragraph{SCAN} \citep{lake_scan} consists of natural language commands that need to be mapped to sequences of actions (e.g. jump twice $\rightarrow$ JUMP JUMP). Among various compositional generalization \textit{splits} of SCAN, we use the Maximum Compound Divergence (MCD) splits introduced by \citet{keysers_measuring_2020}. These splits capture a general notion of compositional generalization by capturing both systematicity and productivity. They maximize the divergence between the compounds while maintaining the closeness of the atom frequency distribution. There are three SCAN-MCD splits with increasing difficulty (i.e., MCD1 being the easiest), each with 8,365 train and 1,045 test examples. 

\paragraph{COGS} \citep{kim_cogs_2020} is a semantic parsing dataset with a diverse set of natural language sentences. The compositional generalization split called "Gen(eralization)" was constructed based on various kinds of linguistic generalizations found in English, such as generalizing the subject role to the object role (systematicity) or generalizing to sentences with more depth (productivity). The training set consists of 24,155 examples while 21,000 examples make up the test.

\paragraph{GeoQuery / GEO} \citep{geoquery, dong-lapata-2016-language} is a semantic parsing dataset consisting of natural language database queries. We use the TMCD compositional split \citep{shaw-etal-2021-compositional} which adapts the MCD principle to a non-synthetic dataset. It is a fairly small dataset consisting of 440 examples for both train and test.

\subsection{Preprocessing}
For SCAN and COGS, we preprocess the output sequences to reduce their lengths in order to be able to fit longer trajectories into memory (i.e., increase $M$). For SCAN, we represent the action sequences in Python syntax as was done in \citet{zhou2023leasttomost} for evaluating LLMs. For example, "LOOK LOOK" is represented as LOOK * 2. For COGS, we omit the brackets and represent the variables $x_n$ as $n$.  Both are intermediate representations that can be fully mapped  back to the original form. For GEO, we follow the same preprocessing step introduced in the original paper for TMCD \citep{shaw-etal-2021-compositional} which replaces the entities with a placeholder. This further brings down the unique number of train examples to 262.
See Appendix \ref{a:preprocessing} for more details.

\subsection{Model Configuration and Training}
We use an 8-layer 8-head causal Transformer with a model dimension of 512 and a feedforward dimension of 2,048 with absolute sinusoidal positional encoding \citep{vaswani}, using the basic implementation available in PyTorch \citep{pytorch}. We do not use any pre-trained weights, and initialize the model and the word embeddings from scratch. 
Appendices \ref{a:hyperparams} and \ref{a:checkpoint} provide details on the used hyperparameters and method of checkpoint selection.

For SCAN, we study the effect of both applying and not applying shuffling of output labels, as its relatively small vocabulary ($|\mathrm{V}|=30$) affords us a full coverage of all words with a few samples. In contrast, COGS and GeoQuery have a much bigger vocabulary ($|\mathrm{V}|=871$ and $|\mathrm{V}|=154$, respectively). Therefore, we do not consider shuffling.

\subsection{Evaluation}
\label{sec:evaluation}

For evaluation, we report \textit{sequence-level} accuracy, where a
sequence is only deemed correct if it is predicted
completely correctly. For each accuracy result, we also report the number of
randomly sampled training examples $k$ used for testing. If not
mentioned, we set $k$ to be one less than the maximum roll-out length
$M$. 
All results report averages over five training runs for SCAN and GEO and three for COGS (for computational reasons).

\subsection{Baselines and Points of Comparison}
\citet{herzig_unlocking_2021} showed that intermediate representations can lead to an improved compositional generalization. This especially applies to SCAN. Hence, we additionally train a 3-layer encoder-decoder Transformer \citep{vaswani} and Universal Transformer \citep{dehghani2019universal} with absolute positional encoding for SCAN. We also train the same baselines for GEO, as they are not reported in the literature. As for COGS, we find
the impact of preprocessing on COGS to be minimal and different than the format \citep{ontanon-etal-2022-making} which appears to be optimal (see Appendix \ref{a:preprocessing} for details). Hence, for COGS, we report the Transformer and Universal Transformer results from the literature.


For all datasets, we also report on a causal Transformer baseline trained using standard supervised learning, which is equivalent to training with the roll-out length of 1 (i.e., $M=1, k=0$). Finally, we compare our approach with the prior meta-learning work: the MAML-augmented Transformer with Tree-based search for COGS and string-based for SCAN \citep{conklin_meta-learning_2021}.

\begin{table*}[tbh]
\centering
\setlength{\tabcolsep}{5.5pt}
\begin{tabular}[tbp]{lcclllll}
\toprule
Method & \multirow{2}{*}{\rotatebox{270}{\hspace*{-1em}Bidir}} & \multirow{2}{*}{\rotatebox{270}{\hspace*{-1em}IntRep}} & \multicolumn{3}{c}{SCAN} & \multicolumn{1}{c}{COGS} & \multicolumn{1}{c}{GEO}\\
  \cmidrule(r){4-6} 
  \cmidrule(lr){7-7} 
  \cmidrule(l){8-8} 
  &  & & \multicolumn{1}{c}{MCD1} & \multicolumn{1}{c}{MCD2}  & \multicolumn{1}{c}{MCD3} &\multicolumn{1}{c}{Gen} & \multicolumn{1}{c}{TMCD}\\
  \midrule

Transformer (lit.)   &  + & - &  \phantom{0}0.4\scriptsize{$\pm$ 0} [1] &  \phantom{0}1.8\scriptsize{$\pm$ 0} [1]   &  \phantom{0}0.5\scriptsize{$\pm$ 0} [1]    &           35\scriptsize{$\pm$ 6} [2]  & NA     \\
Transformer (ours)    &  +  & + &  41.7\scriptsize{$\pm$ 4} &       20.3\scriptsize{$\pm$ 5} &      17.1\scriptsize{$\pm$ 6}  &            80\scriptsize{$\pm$ 0} [3]   & 36.8\scriptsize{$\pm$ 1}  \\
Universal Transformer (ours)    &  +  & + &  36.4\scriptsize{$\pm$ 9} &   34.1\scriptsize{$\pm$ 6}      &    25.5\scriptsize{$\pm$ 10}    &   78\scriptsize{$\pm$ 0} [3]           & 37.3\scriptsize{$\pm$ 1}    \\
Transformer + MAML (lit.)     &  +  & - &  \phantom{0}2.6\scriptsize{$\pm$ 0} [4]   &   \phantom{0}5.6\scriptsize{$\pm$ 1} [4]     &  \phantom{0}6.7\scriptsize{$\pm$ 1} [4]      &    66.7\scriptsize{$\pm$ 4} [4]          & NA      \\
C-Transformer (ours)  & -   & + & 21.8 \scriptsize{$\pm$ 3}  &   25.6\scriptsize{$\pm$ 2}     &    19.7\scriptsize{$\pm$ 2}  &   51.9\scriptsize{$\pm$ 4} & 37.4\scriptsize{$\pm$ 1} \\ \midrule

\textbf{C-Transformer + meta-ICL }  & - & + & 60.4\scriptsize{$\pm$ 13}   &  53.3\scriptsize{$\pm$ 2}  &  \textbf{50.7}\scriptsize{$\pm$ 7}      &    75.7\scriptsize{$\pm$ 1}& \textbf{40.8}\scriptsize{$\pm$ 1} \\
\textbf{C-Transformer + meta-ICL + LB }  & - & + & \textbf{71.2}\scriptsize{$\pm$ 7}   &  \textbf{74.8}\scriptsize{$\pm$ 9}  &  38.7\scriptsize{$\pm$ 8}      &    NA & NA\\\bottomrule

\end{tabular}
\caption{Exp.\ 1: Mean sequence-level accuracies and standard deviations across runs. For our meta-ICL C(ausal)-Transformer, we present the results from best $M$ with $k=M-1$. "Bidir" stands for bidirectional (vs. causal). "IntRep" indicates the use of the optimized intermediate representation for SCAN. Best model on each dataset boldfaced. ``ours'': own experimental results, ``lit.'': results from literature. "LB": label-shuffling. References: [1] \citet{furrer_compositional_2021}, [2] \citet{kim_cogs_2020}, [3] \citet{csordas_devil_2022}, [4] \citet{conklin_meta-learning_2021}.}
\label{tab:main_results}
\end{table*}

\section{Experiments and Results} \label{s:results}

We now present the results of five experiments to better understand the relationship between in-context
learning and compositional generalization: (1) We compare the
performance of our models using $k=M-1$ supports for
evaluation with the baselines. (2) We test the effect of training the model with longer
trajectories (i.e., bigger $M$) which is equivalent to training with a larger number of unique in-context learning problems. 
(3) We test the effect of varying the number of support \textit{training} examples used during
evaluation (i.e., varying $k$) to test whether the models are generalizing through in-context learning. (4) We test how general the in-context learning extracts the latent parameters by testing its ability to learn from a new distribution, providing the model with \textit{test} examples. (5) We test whether a pre-trained model can also benefit from additional training with meta-in-context learning.

\subsection{Exp.\ 1: Main Results}

Table \ref{tab:main_results} summarizes the main results of a causal Transformer trained from scratch using our meta-training method, along with the baselines. 

We first make the observation that our intermediate representation for SCAN leads to an improvement in compositional generalization (compare rows 1 and 2). Although the improvement is substantial, the datasets still remain difficult for the models and the relative difference of difficulty between the MCD splits are retained (see the decreasing performance for SCAN in columns 4-6). We also note that our causal Transformer baseline (row 5) performs mostly worse than the encoder-decoder counterparts, probably due to its unidirectionality. For all three MCD splits, our causal Transformer trained with meta-in-context learning (row 6 and 7) substantially
outperforms all other approaches with or with label shuffling. For COGS, it beats all models except  the Transformer model of \citet{csordas_devil_2022}. For GEO, it seems to provide only a small gain in performance, probably due to its small size which makes it easier to memorize. Next, label shuffling seems to provide a positive boost for the first two splits of SCAN compared to their counterparts. 

It is interesting to note that for the models without label shuffling, the improvement over the causal baseline simply comes from how the data was presented to the model. While it is unclear whether the improvement can be attributed to their in-context learning ability (explored in Exp. 3), we believe that training on linear orderings of examples at least resulted in some form of regularization, where the pressure to learn representations not only for the prediction but also for their use in the future contributed to the improvement.
Since we did not use any informed strategy, such as a retriever, to construct the meta-training distribution, we see these results as evidence for a strong and general inductive bias in-context learning can provide for compositional generalization. 

For the ablation studies (Exp.\ 2--5), we use only SCAN and COGS
since GEO is too small for stable experimentation.



\subsection{Exp.\ 2: More Learning Problems} \label{result:improvement_traj}

\paragraph{Setup.} Next, we investigate how the performance of our model changes when training on trajectories of different lengths. Constructing the task distribution with longer trajectories gives us more unique few-shot learning problems, which we expect to lead to better compositional generalization. However, longer trajectories can also lead to more overfitting, as the model needs to extrapolate less given more support samples. Hence, we train three different values of $M = \{10, 25, 50\}$ for SCAN and $\{5, 10, 25\}$ for COGS, evaluating with $k=M-1$.
\begin{figure}[tb!]
  \centering
  \includegraphics[width=\columnwidth]{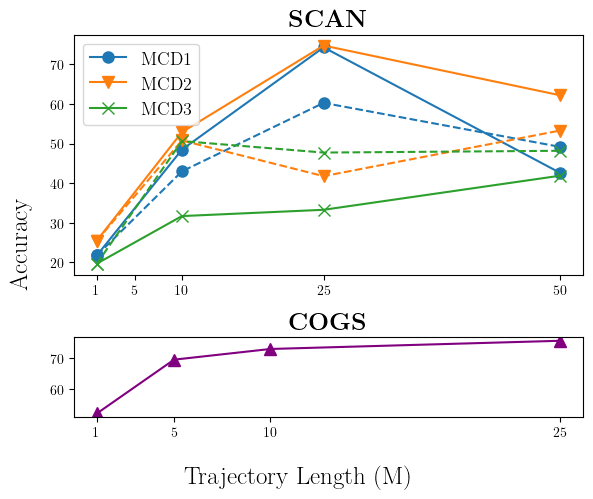}
 \caption{Exp.\ 2: Models trained on different lengths of trajectories (i.e., $M$), with $k=M-1$. $M=1$ is equivalent to the causal Transformer baseline. Dotted lines: models without label shuffling. }
  \label{fig:improvement_longer_trajectories}
\end{figure}

\paragraph{Result.}  The results in Figure \ref{fig:improvement_longer_trajectories} show that an improvement for MCD1 and MCD2 of SCAN from increasing the trajectory length from 10 to 25, but we see signs of overfitting as we increase further, for the label shuffled models. We see a slightly different trend for the non-label-shuffled models for these splits, probably due to memorization, but still $M=25$ and $50$ perform better than $10$. For MCD3, we see a monotonic improvement as we increase the trajectory length for both kinds. For COGS, we see a similar trend with improving performance as the length of the trajectories are increased. The gain diminishes after a certain point because the task becomes easier for the model. Where this turning point occurs depends on the available few-shot learning problems implicit in each dataset; this could be tuned using a standard hyperparameter search. In sum, the in-context learning ability of Transformers is sensitive to the kinds and number of few-shot generalization problems that it is exposed to during training, and having to solve more unique in-context learning problems can lead better compositional generalization.




\begin{figure}[tb!]
  \centering
  \includegraphics[width=\columnwidth]{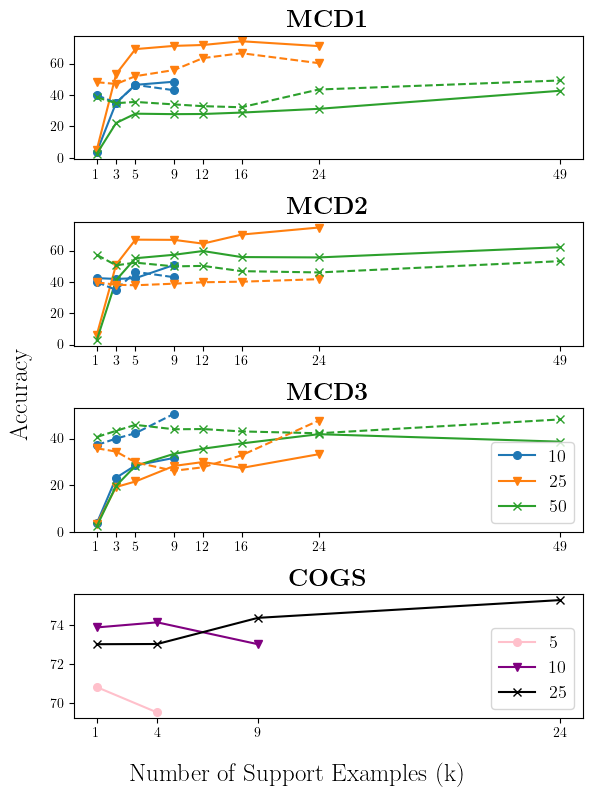}
    \caption{Exp.\ 3: Models evaluated on different numbers of support examples $k$. Lines differ in $M$ (max. roll-out length of meta-training trajectories). Dotted lines: models without label shuffling.  
    }
  \label{fig:varying_k}
\end{figure}

\begin{table*}[t!]
  \centering
\begin{tabular}[tbp]{llllll}
\toprule
& Method & \multicolumn{3}{c}{SCAN} & \multicolumn{1}{c}{COGS} \\
  \cmidrule(r){3-5}
  \cmidrule(l){6-6}
  &&\multicolumn{1}{c}{MCD1} & \multicolumn{1}{c}{MCD2}  & \multicolumn{1}{c}{MCD3} & \multicolumn{1}{c}{Gen}\\
  \midrule
\multirow{2}{*}{Causal Transformer}& baseline & 21.8 \scriptsize{$\pm$ 3}  &   25.6 \scriptsize{$\pm$ 2}     &    19.7 \scriptsize{$\pm$ 2}  &   51.9 \scriptsize{$\pm$ 4} \\
& Meta-ICL & 71.2 \scriptsize{$\pm$ 7}   &  \textbf{74.8} \scriptsize{$\pm$ 9}  &  50.7 \scriptsize{$\pm$ 7}      &    75.7 \scriptsize{$\pm$ 1} \\\midrule
\multirow{2}{*}{GPT-2} & fine-tuning & 42.8 \scriptsize{$\pm$ 15}   &  49.6 \scriptsize{$\pm$ 16}  &  39.6 \scriptsize{$\pm$ 2}      &    81.0 \scriptsize{$\pm$ 0} \\
& Meta-ICL & \textbf{84.7} \scriptsize{$\pm$ 3}   &  71.5 \scriptsize{$\pm$ 10}  &  \textbf{69.1} \scriptsize{$\pm$ 7}      &    \textbf{82.4} \scriptsize{$\pm$ 0} \\\bottomrule
\end{tabular}
\caption{Exp.\ 5: Comparison of causal transformer (Causal Tr., copied from Tab.\ \ref{tab:main_results}) and GPT-2 with and without meta-ICL training, with best $M$ for each dataset (mean sequence-level accuracies and standard deviations across runs).}
\label{tab:GPT2}
\end{table*}

\subsection{Exp.\ 3: Number of Demonstrations} \label{result:vary_k}
\paragraph{Setup.} So far, we have only used $k=M-1$ support
examples for evaluation. In this experiment, we investigate how the model
generalizes for different number of support examples. If the models are truly generalizing through in-context learning, 
then it should perform better with more demonstrations (i.e., larger $k$). If this is the case, then we also wish to rule out whether the improvement from Exp. 2 was
simply due to the models with higher $M$ having access to more
support examples during evaluation by comparing with the same $k$. To do so, we evaluate
the SCAN-MCD models using different values of $k=\{1, 3, 5, 9, 12, 16,
24, 49\}$ using the full test set. For COGS, we take 20\% of the test
set and vary $k=\{1, 4, 9, 24\}$\footnote{The reason is that the test set of COGS is 21 times larger than SCAN. We found the results on our sample to be
representative for evaluating on the entire test set.}. Note that we only evaluate when $k$ does not
exceed the maximum roll-out length for each given model (i.e., $k < M$).
%

%
%
\paragraph{Result.} Figure \ref{fig:varying_k} shows the
results. For SCAN, all models improve as it receives more and more
support examples, even for the models trained without label shuffling. The result also confirms the conclusion from Exp.\ 2: For all MCD splits of
SCAN, even when the model is given the same number of support
examples, the best model performs better than the rest (i.e., 25 for
MCD1 and MCD2 and 50 for MCD3 with label shuffling), suggesting that these models did learn more generalizable in-context learning.

For COGS we do not see a clear trend, except for $M=25$. We believe that this can be attributed to the \textit{informativeness} of support examples.  We non-rigorously define \textit{informativeness} as the set of underlying latent rules that governs the input-output mappings available in the given context relevant for answering the next (i.e., test) example.  Since no shuffling was applied, the COGS models appear to have memorized the examples, though it was being regularized in doing so. 
The pressure to memorize seems to be stronger in COGS because the dataset is more diverse
than SCAN, hence the support examples that make up the context are less likely to be
informative. When trained with larger $M$ (e.g., $M=25$), where
informative examples become more probable, the COGS model shows a
similar pattern to SCAN, where the past examples are bound to be more
informative due to its small vocabulary. This bolsters the importance
 of context informativeness for the emergence of in-context
learning in Transformers.
This result is also consistent with the findings of \citet{chan_data_2022} who demonstrated that informative context can drive the emergence of in-context learning, even despite the possibility of memorization.




\subsection{Exp.\ 4: New Distributions}

\paragraph{Setup.} We now test how general our models can in-context learn by experimenting how well they can learn from a new distribution. For SCAN splits, we hold out 49 examples from the \textit{test} set to sample our support examples during evaluation. We chose the value 49 as this is the maximum $k$ for any model. We then evaluate our models on the rest of the test set by sampling from the held out \textit{test} examples to form the contexts. We repeat Exp.\ 3 and report on the relative improvement (RI), $\frac{new-old}{old}\times 100 $.
\begin{figure}[tb!]
  \centering
  \includegraphics[width=\columnwidth]{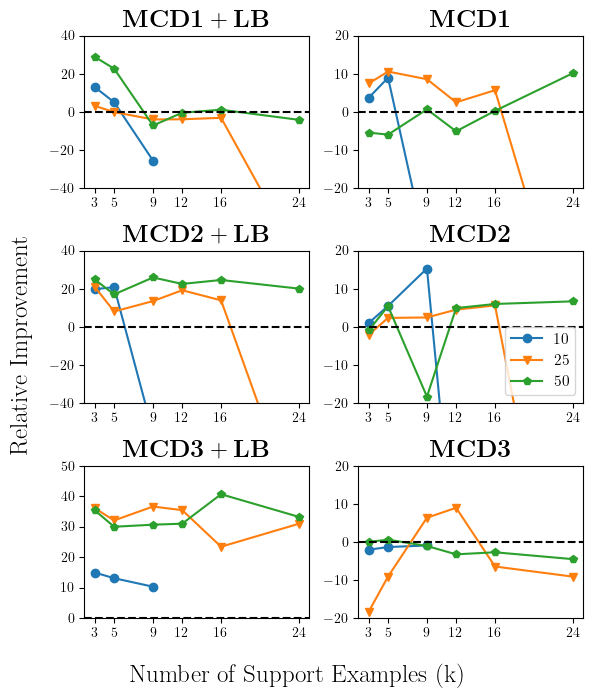}
  \caption{Exp. 4: Models evaluated with different numbers of support examples $k$ sampled from the held-out portion of the \textit{test} set for both with label shuffling (LB) (left column) and without (right). Relative improvement (RI) is calculated using $\frac{new-old}{old}\times 100 $.}
  \label{fig:varying_k_test}
\end{figure}


%
\paragraph{Result.} As Figure
\ref{fig:varying_k_test} shows, for every label-shuffled model (left column), there exists a value of $k$
that leads to an improved performance. This suggests that the learned 
in-context learning ability is \textit{general} to a degree, being
able to learn from \textit{test} examples. This is especially true for
MCD2 and MCD3, where the model improves for most (MCD2) and all (MCD 3) values of $k$. This is interesting, because these are the more difficult
splits of SCAN, where the \textit{test} support examples can be much
more informative. Contrarily, the models without label-shuffling (right column) do not show clearly an ability to 
learn from a novel distribution, especially for the more difficult splits of SCAN where the label-shuffled counterparts benefited the most. This demonstrates the existence of conflict between memorization and in-context learning in Transformers, where the possibility of memorization can negatively affect their ability to compositionally generalize using in-context learning upon its emergence. 

For clarity, Figure \ref{fig:varying_k_test} omits $k$=1 and
$k=49$. $k=1$ leads to a significant positive RI for the LB models,
and $k=49$ leads to significant deterioration for all models by
failing to predict the end-of-sequence token properly (omitted for clarity). This is probably due to the model
having to predict in positions far exceeding the maximum value of
position seen during training, which has been known to be difficult
for Transformers \citep{newman-etal-2020-eos, csordas_devil_2022}.

\subsection{Exp. 5: Meta-ICL for Pre-trained Models}

\paragraph{Setup.} Finally, we investigate how the effects of applying the meta-ICL
framework carry over from models trained from scratch to the type of
pre-trained causal language model typically used in NLP such as GPT-2 \citep{radford2019language}.
We train two types of GPT-2 models without label shuffling for SCAN and COGS\footnote{We use Huggingface "gpt2" \citep{wolf-etal-2020-transformers} }: a supervised fine-tuned model ($M=1$) and meta-ICL models $M=\{10, 25\}$.  Details on training, hyperparameters, and checkpointing are provided in Appendices \ref{a:hyperparams} and \ref{a:checkpoint}.

\paragraph{Result.} Table \ref{tab:GPT2} summarizes the
result. Evidently, pre-trained models such as GPT-2 can also benefit
from further training with meta-in context learning (compare rows 3
and 4). GPT-2 also outperforms the models trained from scratch with
meta-ICL (row 2) in 3 out of 4 settings. However, the performance gain
from meta-ICL training is diminished due to the pre-training (compare
rows 3 and 4). In sum, these results confirm that pre-training on a
large collection of natural language data lead to an acquisition of a
prior more conducive to in-context learning, which agrees with the
general consensus in the research community \citep{brown2020language,
  chowdhery2022palm, min-etal-2022-metaicl}. Hence, our results are
also relevant for the types of pretrained Transformer-based causal
language models in widespread use in NLP today.



\section{Conclusion} \label{s:conclusion}

In this paper, we have studied the emergence 
of compositional generalization with a meta-learning regime which forces a sequence-to-sequence model to learn to \textit{in-context learn}. We have investigated on three difficult datasets: the MCD splits of SCAN, COGS, and TMCD of GEO. Our main results showed that the meta-trained models show substantially better compositional generalization than the baselines in SCAN and GEO and closely matching in COGS. 


Our results provide evidence that in-context learning can induce
compositional generalization. We confirm this relationship through
various ablative studies, illustrating that compositional
generalization can be improved by training on more in-context
learning problems, in-context learning can emerge when learning from informative
contexts despite the possibility of memorization, and the existence of
conflict between memorization and compositional generalization.  In
this way, our study represents one step towards a deeper
understanding of in-context learning. This can arguably improve our
handling of out-of-distribution generalization, which is a fundamental
challenge for effective machine learning
\citep{Ye_Zhu_Wang_Zeng_Shao_Peng_Pan_Li_Zhu_2023}, as well as making
our learning models more plausible on the cognitive side, given the
very limited memorization capabilities of humans
\citep{fodor_connectionism_nodate}.

In future work, we plan to investigate the effect of using relative positional encoding \citep{dai_transformer-xl_2019} which are widely applied in recent LLMs as it has been shown to improve compositional generalization \citep{ontanon-etal-2022-making}.
Second, we believe that it would be worthwhile to
investigate the effect of equipping the model with retrieval method to better choose the support examples in the future.






\section*{Limitations}
Our results are relevant for generally elucidating the relationship between in-context learning and compositional generalization for a class of learning algorithms with memory. However, it must be observed that our work is  predominantly empirical in nature and is not supported by theoretical guarantees. Also, it is conducted in an experimental lab-setting utilizing compositional generalization benchmarks that are well-established but are of synthetic nature. Hence, there might exists exceptions to the presented observations. The practical applications of the proposed methods for larger real world datasets are uncertain and need to be observed.

Furthermore, our approach in assessing a model’s compositional generalization follows the field's predominant approach to this question: by generating a single data distribution and splitting them into a train and test in a systematic manner to test for compositionality. Although we have experimented with multiple datasets, there are still limitations in each chosen method of splitting used in our work. This implies that the degree of learned compositionality of our models might only partially encompass the entire spectrum of compositionality present in human languages.

\section*{Ethics Statement} \label{s:limit_future}

Our work is concerned with foundational questions of learning generalizable models. It does not introduce new risks, nor does it involve sensitive applications. The datasets and pre-trained models are publicly available. Computational costs for training are relatively low. We do not believe that there are substantial ethical concerns in our work.


\section{References}
\label{sec:reference}

\bibliography{custom, thesis, anthology}

\newpage
\appendix

\section{Hyperparameters and Computing Resource} \label{a:hyperparams}
\paragraph{Causal Transformer} We use the PyTorch \citep{pytorch} implementation of RAdam \citep{Liu2020On} as our choice of optimizer with the learning rate of $1\times10^{-4}$ and $\beta=(0.9,0.99)$ for all of our experiments. For stable training, we apply a linear warm-up for 500 steps for SCAN and GeoQuery and 5000 for COGS. We clip the gradient whenever the norm exceeds 5. We apply dropout rate of 0.1, ReLU activation, and  batch-size of 5. Small batch size was used due to the limited availability of computing resources. Parameters are initialized according to the default initialization method of PyTorch. This means that the word embeddings are initialized by drawing from a standard normal. The embeddings have the same dimension as the model. We use the variant where the LayerNorm \citep{ba2016layer} is applied before each sub-block for SCAN and a normal configuration for COGS and GeoQuery. The resulting model size has 25.2 million parameters for all datasets. For the causal baseline (i.e., $M=1$),  use of a low batch-size leads to unstable training, hence we increase the batch-size to 256. Finally, we note that we did not perform any systematic hyperparameter tuning and most of the used hyperparameters were initial guesses.

\paragraph{Transformer and Universal Transformer} Both Transformer and Universal Transformer baselines are a 3-layer encoder-decoder architecture, and it is adapted from the code release of \citet{csordas_devil_2022}. We use the same learning rate $1\times10^{-4}$ and $\beta=(0.9,0.99)$ using Adam \citep{kingma2014adam} as our optimizer of choice, We do not use the default learning rate value of PyTorch as we saw the alternative configuration to be more stable. The dimension of 128 is used for both model state and word embeddings with 8 heads and 256 for feed-forward dimension. We use the dropout rate of 0.1 and batch size of 256. 

\paragraph{GPT-2} We use the same set of hyperparameters as the causal Transformer models trained from scratch described above, except for the learning rate which we set to $5 \times 10^{-5}$.

\paragraph{Computing Resources} We used a single GeForce RTX 2080 Ti 11G for our SCAN and GeoQeury experiments and a single GeForce GTX TITAN X 12G for our COGS experiments. 

\section{Checkpoint Selection for Evaluation} \label{a:checkpoint}
For SCAN, we follow the checkpoint selection method of \citet{conklin_meta-learning_2021} and use the available development set to pick the checkpoint for testing by training for 20k steps evaluating every 1000 steps. Usually, each model with meta-training takes around 10k steps to converge. The causal Transformer baselines, Transformers, and Universal Transformers all take much less time to converge, hence we only train for 10k steps.

For COGS, we simply train the models for 150k steps and take the last checkpoint for evaluation. This was similarly done in \citet{conklin_meta-learning_2021}, but they use 10\% of the test set to tune their hyperparameters. There is a validation set associated with the training set in COGS, but it is a widely known that tuning on this set does not work well \citep{csordas_devil_2022} as the model continues to improve on the test even when the model scores perfectly on the train and validation set. 

For GeoQuery, we train the models for 50k steps and select the last checkpoint for evaluation. 

\section{Datasets and Preprocessing} \label{a:preprocessing}
\paragraph{SCAN} The preprocessing decreases the average output length of the dataset from 14.3 to 12.2 and the maximum sequence length from 48 to 17. In the new format, the overall vocabulary size of SCAN is 30 with 11 output words, 4 special symbols and 15 input words. Table \ref{table_SCAN_examples} shows a few examples. 

\paragraph{COGS} The resulting preprocessing is illustrated in Table \ref{table_COGS_examples}. Before preprocessing, the average length is 51.07 and maximum of 175 which becomes 28.01 and 96 respectively after preprocessing. The resulting vocabulary size is 871. 

\paragraph{GeoQuery} We do not apply any pre-processing for intermediate representation. As mentioned in the main text, we only replace the entities with placeholders. 

We noted that our intermediate representation method is different from the format found to be useful, which converts the sequence prediction task to a sequence tagging task \citep{ontanon-etal-2022-making}. This is possible because in COGS, the target output is a concatenation of five semantically parsed "tags": a parent, the role of the parental relation, the category, the noun determiner, and the verb name. Hence, instead of generating the output sequence, a model can be made to tag each input token in parallel and combine the results for the final prediction.

\begin{table*}[!htbp]
 \centering
\begin{tabularx}{\linewidth}{CCC}
\toprule
 \textbf{Command (Input)}                   & \textbf{Before (Output)}          & \textbf{After (Output)}                                                                            \\ \cline{1-3}
 run twice              & RUN RUN           & RUN * 2                                                                                        \\\hline
jump after run    & RUN JUMP                & RUN  + JUMP                                                                                                    \\\hline
jump around right       &  RTURN JUMP RTURN JUMP RTURN JUMP RTURN JUMP        &  ( RTURN JUMP ) * 4                              \\\hline
jump around right and walk twice & RTURN JUMP RTURN JUMP RTURN JUMP RTURN JUMP WALK WALK & ( RTURN JUMP ) * 4 + WALK * 2   \\ \hline
\end{tabularx}

\caption[SCAN examples using Python syntax]{Example SCAN action sequences (outputs) before and after preprocessing.}
\label{table_SCAN_examples}
\end{table*}

\begin{table*}[!htbp]
 \centering
\begin{tabularx}{\linewidth}{CCC}
\toprule
 \textbf{Sentence (Input)}                   & \textbf{Before (Output)}          & \textbf{After (Output)}                                                                            \\ \cline{1-3}
A rose was helped by a dog .              & rose ( $x _ 1$ ) AND help . theme ( $x _ 3$ , $x _ 1$ ) AND help . agent ( $x _ 3$ , $x _ 6$ ) AND dog ( $x _ 6$ )           &  rose 1 AND help . theme 3 1 AND help . agent 3 6 AND dog 6    
\\\hline
Charlie loaned the cake in a house to the girl . & * cake ( $x _ 3$ ) ; * girl ( $x _ 9$ ) ; loan . agent ( $x _ 1$ , Charlie ) AND loan . theme ( $x _ 1$ , $x _ 3$ ) AND loan . recipient ( $x _ 1$ , $x _ 9$ ) AND cake . nmod . in ( $x _ 3$ , $x _ 6$ ) AND house ( $x _ 6$ ) & * cake 3 ; * girl 9 ; loan . agent 1 Charlie AND loan . theme 1 3 AND loan . recipient 1 9 AND cake . nmod . in 3 6 AND house 6 
  \\ \hline
\end{tabularx}

\caption{Example COGS semantic parsing results before and after preprocessing.}
\label{table_COGS_examples}
\end{table*}


\end{document}